# Data-Centric Learning Framework for Real-Time Detection of Aiming Beam in Fluorescence Lifetime Imaging Guided Surgery

Mohamed Abul Hassan, *Member, IEEE, Pu Sun*, Xiangnan Zhou, Lisanne Kraft, Kelsey T Hadfield, Katjana Ehrlich, Jinyi Qi, Fellow, IEEE, Andrew Birkeland and Laura Marcu, *Senior Member, IEEE*

*Abstract*— This study introduces a novel data-centric approach to improve real-time surgical guidance using fiber-based fluorescence lifetime imaging (FLIm). A key aspect of the methodology is the accurate detection of the aiming beam, which is essential for localizing points used to map FLIm measurements onto the tissue region within the surgical field. The primary challenge arises from the complex and variable conditions encountered in the surgical environment, particularly in Transoral Robotic Surgery (TORS). Uneven illumination in the surgical field can cause reflections, reduce contrast, and results in inconsistent color representation, further complicating aiming beam detection. To overcome these challenges, an instance segmentation model was developed using a data-centric training strategy that improves accuracy by minimizing label noise and enhancing detection robustness. The model was evaluated on a dataset comprising 40 *in vivo* surgical videos, demonstrating a median detection rate of 85%. This performance was maintained when the model was integrated in a clinical system, achieving a similar detection rate of 85% during TORS procedures conducted in patients. The system's computational efficiency, measured at approximately 24 frames per second (FPS), was sufficient for real-time surgical guidance. This study enhances the reliability of FLIm-based aiming beam detection in complex surgical environments, advancing the feasibility of real-time, image-guided interventions for improved surgical precision.

## I. Introduction

FIBER-BASED fluorescence lifetime imaging (FLIm) has demonstrated great potential as a tool for providing intraoperative guidance, especially during tumor resection. In transoral robotic surgery (TORS), the flexible fiber optic probe allows for freehand scanning and can be easily integrated with minimally invasive robotic systems, such as the Da Vinci surgical system [1], [2]. This integration allows surgeons to visualize tumor regions in the oropharynx that are difficult to detect using only visual inspection and tactile feedback. The intricate anatomy of the oropharyngeal region often restricts the effectiveness of rigid, forward-viewing endoscopes [3], [4] or handheld camera-based FLIm systems [5], emphasizing the growing importance of flexible, minimally invasive probes for enhanced for surgical guidance.

In addition, the use of fiber-based FLIm extends beyond the oropharynx to include surgeries in the gastrointestinal tract [6], brain [7], and lungs [8], demonstrating its versatility across various cancer procedures. Freehand scanning with these fiber probes allows surgeons to collect point measurements across the tissue surface, providing greater control over the imaging process [9]. However, accurately localizing and coregistering these measurements for visualization on the tissue surface remains a considerable challenge. To aid in this, an aiming beam (440 nm) generated by a continuous-wave laser is integrated with the pulsed fluorescence excitation light within a single delivery and collection fiber. This visible aiming beam allows surgeons to pinpoint the exact location of FLIm measurements in real-time. Despite this integration, factors such as uneven illumination in the surgical field can introduce reflections, reduce contrast, and lead to inconsistent color representation, complicating aiming beam detection and precise FLIm measurement localization.

Accurate detection of the aiming beam is an essential first step in visualizing the interrogated tissue region. Although deep learning algorithms can achieve near-perfect accuracy and real-time performance—often surpassing human abilities when

This work was supported by the National Institutes of Health under Grant 2R01CA187427 in collaboration with Intuitive Surgical, Inc; and P41 -EB032840-01. All authors are affiliated with the *University of California, Davis*. Mohamed Abul Hassan, *Pu Sun*, Xiangnan Zhou, Lisanne Kraft, Kelsey T Hadfield, Katjana Ehrlich, Jinyi Qi, work within the *Department of Biomedical Engineering*. Andrew C. Birkeland practice oncologic surgery within the *Department of Otolaryngology – Head & Neck Surgery*. Laura Marcu work within the *Department of Neurological Surgery and the Department of Biomedical Engineering* Corresponding Author Laura Marcu (e-mail: lmarcu@ucdavis.edu). Ethical Considerations: All procedures performed in studies involving human participants were in accordance with the ethical standards of the institutional research committee and with the 1964 Helsinki Declaration and its later amendments or comparable ethical standards. Ethical approval for the study was obtained from the University of California, Davis Institutional Review Board (IRB), and informed consent was obtained from all individual participants included in the study. The use of in vivo surgical videos and patient-level data was conducted under strict adherence to these ethical guidelines, ensuring patient confidentiality and data protection throughout the study.

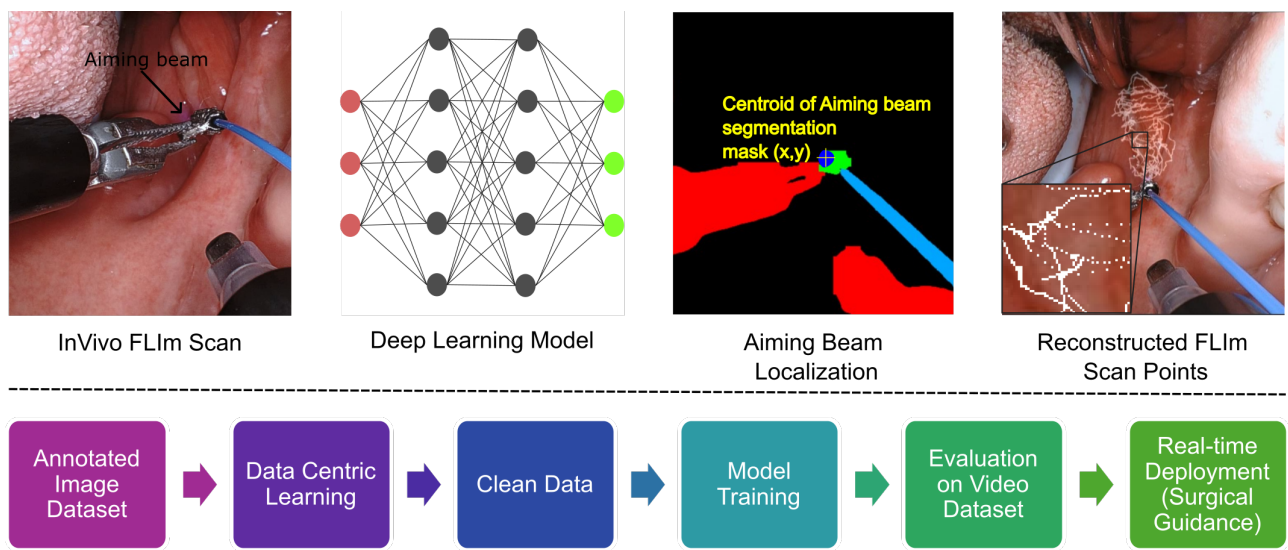

Figure 1: Overview of the proposed methodology for real-time aiming beam detection using a deep learning model. (upper panel) Illustrates the point-level FLIm scan over the in vivo tissue surface. The trained deep learning model performs instance segmentation to localize the aiming beam, to reconstruct the FLIm points throughout the scan. (lower panel) Shows the model development pipeline, starting with an annotated image dataset. A data-centric learning approach is applied to identify and remove mislabeled or noisy data, yielding a clean dataset for training. The cleaned data is used to train a Yolov8-n model, which is then evaluated on an independent video dataset to assess its accuracy and robustness.

trained on clear, well-curated datasets—translating these models into clinical practice remains challenging [10]. The availability of annotated data is limited due to complexity of data collection, and patient data is highly heterogeneous. Insufficient data representation can introduce noise and increase the likelihood of the model learning spurious correlations, which diminishes its effectiveness in real-world scenarios [11], [12].

In this study, we present a method that utilizes a deep learning model for detecting the aiming beam addressing the challenge of limited annotated data by employing domain adaptation techniques with a pre-trained model. Additionally, we incorporated a data-centric learning mechanism that handles data heterogeneity by identifying and removing mislabeled data, thereby reducing noise and minimizing the risk of the model learning spurious correlations (see Fig. 1). This approach ensures that the model is trained exclusively on high-confidence, correctly labeled examples, reducing the risk of overfitting and enabling more reliable predictions when deployed deployment in diverse and complex clinical environments.

## II. METHODS AND MATERIAL

### A. FLIm Hardware and Data Acquisition

This study utilized a multispectral FLIm device employing a pulse-sampling technique to acquire data [13]. The device is equipped with a 355 nm UV laser for fluorescence excitation, pulsed at a 480 Hz repetition rate. Excitation light is delivered to the tissue via a 365 μm multimode optical fiber (0.22 NA), which also relays the resulting fluorescence signal to a set of dichroic mirrors and bandpass filters to spectrally resolve the autofluorescence. The autofluorescence is then detected by three UV-enhanced Si APD modules with integrated trans-impedance amplifiers, resolving the signal in the following spectral channels primarily associated with the autofluorescence of collagen (390/40 nm), Nicotinamide Adenine Dinucleotide (NADH) (470/28 nm), and Flavin Adenine Dinucleotide (FAD) (542/50 nm) [14].

The FLIm device incorporates a 440 nm continuous wave laser that serves as an aiming beam. This aiming beam provides real-time visualization of the locations where FLIm point measurements are collected by generating visible blue light illumination at the precise location where data is acquired [15].

FLIm data was collected during two distinct surgical situations: (1) Transoral robotic surgical procedures (TORS) using the da Vinci SP for oropharyngeal cancer, and (2) non-TORS procedures performed manually for oral cavity cancer. The average FLIm scan duration was approximately 45 seconds, resulting in approximately 1,350 averaged point measurements per surgical field scanned. For non-TORS cases, a hand-held fiber probe (Omniguide Laser Handpiece) was used in combination with a endoscopic camera (Stryker). Whereas for TORS cases, the fiber optic probe was actuated by the robotic instruments, with the surgical field visualized through the integrated da Vinci camera.

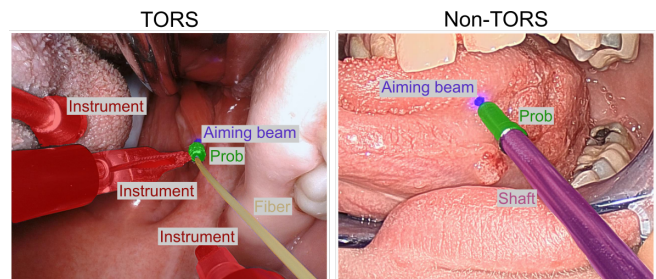

Figure 2: Instance Segmentation Annotation Masks for Surgical Scenes in TORS and Non-TORS Procedures. The figure displays the annotated segmentation masks for various objects within the surgical field, including the aiming beam, surgical instruments, probe, fiber, and shaft. The left panel represents a TORS procedure. The right panel shows a non-TORS procedure.

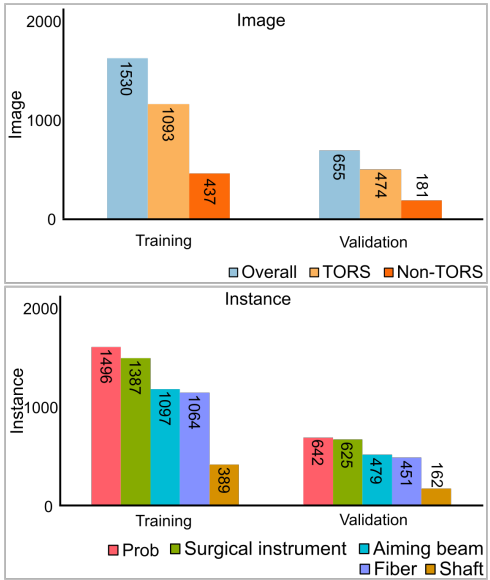

Figure 3: Distribution of images and annotated instances within the training and validation sets. The bar chart illustrates the quantity of images (upper panel) and the number of annotated instances (lower panel) for each category across both datasets used during model training and validation phase.

For the TORS procedures, a fiber probe was used, and a 3D-printed stainless steel grasper tab was added to the distal end of the probe to allow it to be securely grasped and maneuvered by via the da Vinci robotic instruments.

### B. Instance Segmentation

An instance segmentation approach [16] was adopted to accurately segment multiple objects within the surgical scene, including the aiming beam, surgical instruments, and the FLIm probe (see Fig. 2). By generating segmentation masks for these objects, the exact location of the aiming beam's centroid can be determined, this is critical for point tracking and visualization. Additionally, the segmentation masks for surgical instruments, probes, and fibers allow for the exclusion of regions that should not be included in the motion correction process, reducing potential errors and improving the reliability of real-time overlay visualization [9]. The Yolov8n-seg model was chosen for its state-of-the-art performance in instance segmentation [17]. This model is optimized to manage multiple instances with high accuracy and computational speed, making it ideal for real-time surgical applications.

### C. Annotation for Instance Segmentation task

Eight annotators each manually labeled approximately 270 images, utilizing pre-trained models to semi-automatically create polygon masks with the Anylabeling open-source software [18]. Two annotators then reviewed the annotations by overlaying the polygons on the white light images. Any poorly covered annotations were reannotated to ensure high quality. The annotations covered five key instances in the surgical view of the head and neck procedure: surgical instrument, probe, aiming beam, fiber, and shaft, resulting in an annotated dataset of 2185 surgical view images (see Fig. 3).

Table I: Summary of the computational environment used across different stages of model development and deployment.

| Stage | Environment | GPU |
|---|---|---|
| **Training/ Validation** | Python | NVIDIA A100 |
| **Testing** | Python | NVIDIA 2080Ti |
| **Deployment** | C++ | NVIDIA 4070Ti |

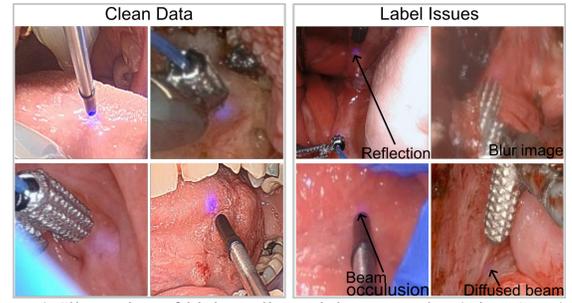

Figure 4: Illustration of high-quality training examples (Clean Data) and noisy training examples (Label Issues) as identified by the data-centric learning approach. The left panel shows well-annotated, clear images used in training, while the right panel highlights examples with annotation errors or inconsistencies that were flagged and filtered out to improve model performance.

### D. Data Centric Learning

The data-centric learning approach improves the model's generalization by prioritizing high-quality data. [19]. This is achieved by identifying mislabeled or noisy data through a confidence scoring mechanism, which removes incorrect and weak annotations from the training dataset. The dataset is represented as: $D = \{(x_i, y_i)\}_{i=1}^{N}$ where ($x_i \in \mathbb{R}^d$) denotes the input image, and $y_i$ denotes the corresponding class labels for object mask. The model is defined as:

$$\hat{y}_i = f(x_i; \theta)$$

where $\theta$ represents the model parameters, and $\hat{y}_i$ is the predicted output, including class probabilities. The confidence score $\sigma_i$ is derived from the class probability. For each segmented object, the model produces a class probability. The confidence score for a designated class $d$ in this case, 'Aiming Beam,' is given by:

$$\sigma_i = P(y = d | x_i; \theta).$$

To create a noise-free training dataset, we filter out samples with low confidence scores below a threshold $\kappa$ and samples where there is no corresponding label for a segmented object, resulting in a cleaned dataset without incorrect and weak annotations $D_{clean} = \{(x_i, y_i) \in D \mid \sigma_i \geq \kappa \text{ or } y_i = \hat{y}_i\}$.

### E. Model Training, Validation and Testing

*Training and Validation*: We trained the YOLOv8n-seg model, a segmentation variant of the YOLOv8 architecture, to perform instance segmentation on a curated on a curated dataset of head and neck surgical view images. The model architecture consists of 261 layers and approximately 3.26 million trainable parameters, requiring 12.1 GFLOPs for computation. Its backbone includes convolutional layers (Conv) and C2f blocks for efficient feature extraction, along with a Spatial Pyramid Pooling Fast (SPPF) layer to enhance the receptive fields. The segmentation head combines features from multiple layers through concatenation and upsampling operations, enabling accurate and comprehensive instance segmentation.

Pretraining was conducted by transferring weights from the YOLOv8 model trained on the COCO dataset [20]. During training, we froze the model.22.dfl.conv.weight layer to prevent updates to its weights.

Table II. Comparison of validation performance metrics (mAP50 and mAP50-95). Mean average precision at an IoU threshold of 50% (mAP50) and mean Average Precision across IoU thresholds ranging from 50% to 95% (mAP50-95)

| Version | mAP50 | mAP50-95 |
|---|---|---|
| Model v3 | 0.868 | 0.605 |
| Model v3 DC | 0.877 | 0.616 |

Training was performed on 70% of the dataset (1,530 images) over 100 epochs, with a batch size of 16 and an image resolution of 640x640 pixels, using an NVIDIA A100 GPU (see Table. I). The AdamW optimizer, automatically configured with an initial learning rate of 0.00143, was used alongside a linear learning rate schedule. To enhance computational efficiency, Automatic Mixed Precision (AMP) was utilized. Various data augmentation techniques, such as random horizontal flipping, mosaic augmentation, random erasing, rotation, scaling, and translation were applied to improve model generalization. Post-training, the model was validated on the remaining 30% of the dataset (655 images). The best-performing model was selected based on the mean Average Precision (mAP) across different Intersection over Union (IoU) thresholds for all classes see Table. II.

*Testing*: The primary objective of this study was to develop a robust aiming beam detection model to improve real-time visualization for surgical guidance. To evaluate the model's effectiveness, we tested it on an independent and fixed dataset comprising of 40 videos from patients undergoing head and neck surgery. In this dataset, the aiming beam was annotated with center point coordinates corresponding to the 'blue light' of the beam. We assessed the model's accuracy by comparing these annotated coordinates with the centroid coordinates of the aiming beam mask predicted by the model, measuring its performance against that of a human annotator.

### F. Model Deployment in Clinical System

The real-time implementation on the clinical system was achieved by deploying the best-performing model using ONNX Runtime. Inferences were executed on input frames during scans using an OpenCV environment in C++. The model was configured with an input size of 480 by 480 pixels and a class threshold of 0.01. To optimize inference speed, the system leveraged an NVIDIA RTX 4070 Ti GPU, ensuring efficient real-time processing during clinical procedures.

### G. Evaluation

Given the application focus of this study, the primary evaluation metrics were the detection rate and precision of the aiming beam localization. The detection rate was assessed by comparing the model's predictions for the aiming beam in each frame of the video dataset against the human annotations. The precision of aiming beam localization was evaluated by calculating the Euclidean distance and Pearson correlation coefficients between the x and y coordinates ($PCC_x$, $PCC_y$) of the predicted and annotated aiming beam points from the subset of successfully detected instances. Additionally, we evaluated the impact of progressively increasing the training dataset size by incrementally adding more data to the training set.

$$\text{Detection Rate} = \frac{N_{\text{detected}}}{N_{annotated}}$$

$$\text{Euclidean Distance} = \sqrt{(x_p - x_a)^2 + (y_p - y_a)^2}$$

$$PCC_x = \frac{\sum_{i=1}^{N_{\text{detected}}}(x_{p_i} - \overline{x_p})(x_{a_i} - \overline{x_a})}{\sqrt{\sum_{i=1}^{N_{\text{detected}}}(x_{p_i} - \overline{x_p})^2 \sum_{i=1}^{N_{\text{detected}}}(x_{a_i} - \overline{x_a})^2}}$$

$$PCC_y = \frac{\sum_{i=1}^{N_{\text{detected}}}(y_{p_i} - \overline{y_p})(y_{a_i} - \overline{y_a})}{\sqrt{\sum_{i=1}^{N_{\text{detected}}}(y_{p_i} - \overline{y_p})^2 \sum_{i=1}^{N_{\text{detected}}}(y_{a_i} - \overline{y_a})^2}}$$

Where $(\overline{x_p}, \overline{y_p})$ denote the mean x and y coordinates for the predicted points across all detected frames, and $(\overline{x_a}, \overline{y_a})$ denote the mean x and y coordinates for the annotated points. Where, $(x_{p_i}, y_{p_i})$ are the predicted coordinates for the *i*-th detected frame, while $(x_{a_i}, y_{a_i})$ are the annotated coordinates for the *i*-th detected frame. $N_{\text{detected}}$ and $N_{annotated}$ represent the number of frames in which the aiming beam was detected and annotated, respectively.

## III. RESULTS

We examined four model versions: model v1 (trained with 40% of the training data), model v2 (trained with 60% of the training data), model v3 (trained with 100% of the training data), and model v3 DC (a data-centric version of model v3). As shown in Fig. 5, model v3 DC achieved the highest median detection rate at approximately 0.85, with a mean detection rate of 0.78 and a standard deviation of 0.18. This suggests that incorporating data-centric strategies in model v3 DC improved its ability to consistently detect the aiming beam across video frames. Model v3, trained with the full dataset but without the data-centric approach, followed with a mean detection rate of 0.74, showing that while increasing the dataset size enhances performance, the data-centric approach further optimizes it. Model v2 and model v1, trained with 60% and 40% of the data, respectively, demonstrated lower detection rates, with mean values around 0.69, indicating the importance of dataset size in achieving higher detection rates.

Table III: Comparative analysis of the Pearson correlation coefficients for 'x' and 'y' coordinates across different models and surgery types on the test dataset.

| Version | Surgery Type | $PCC_x$ Mean(Std) | $PCC_x$ Median | $PCC_y$ Mean(Std) | $PCC_y$ Median |
|---|---|---|---|---|---|
| model v1 | Overall | 0.991(0.014) | 0.995 | 0.990(0.021) | 0.996 |
| model v2 | Overall | 0.992(0.019) | 0.996 | 0.994(0.006) | 0.996 |
| model v3 | Overall | 0.974(0.100) | 0.994 | 0.972(0.136) | 0.996 |
| model v3 DC | Overall | 0.986(0.023) | 0.995 | 0.991(0.015) | 0.996 |
| model v3 | non-TORS | 0.955(0.153) | 0.997 | 0.943(0.209) | 0.996 |
| model v3 DC | non-TORS | 0.987(0.025) | 0.997 | 0.989(0.016) | 0.994 |
| model v3 | TORS | 0.989(0.012) | 0.994 | 0.994(0.007) | 0.997 |
| model v3 DC | TORS | 0.986(0.022) | 0.993 | 0.992(0.014) | 0.996 |

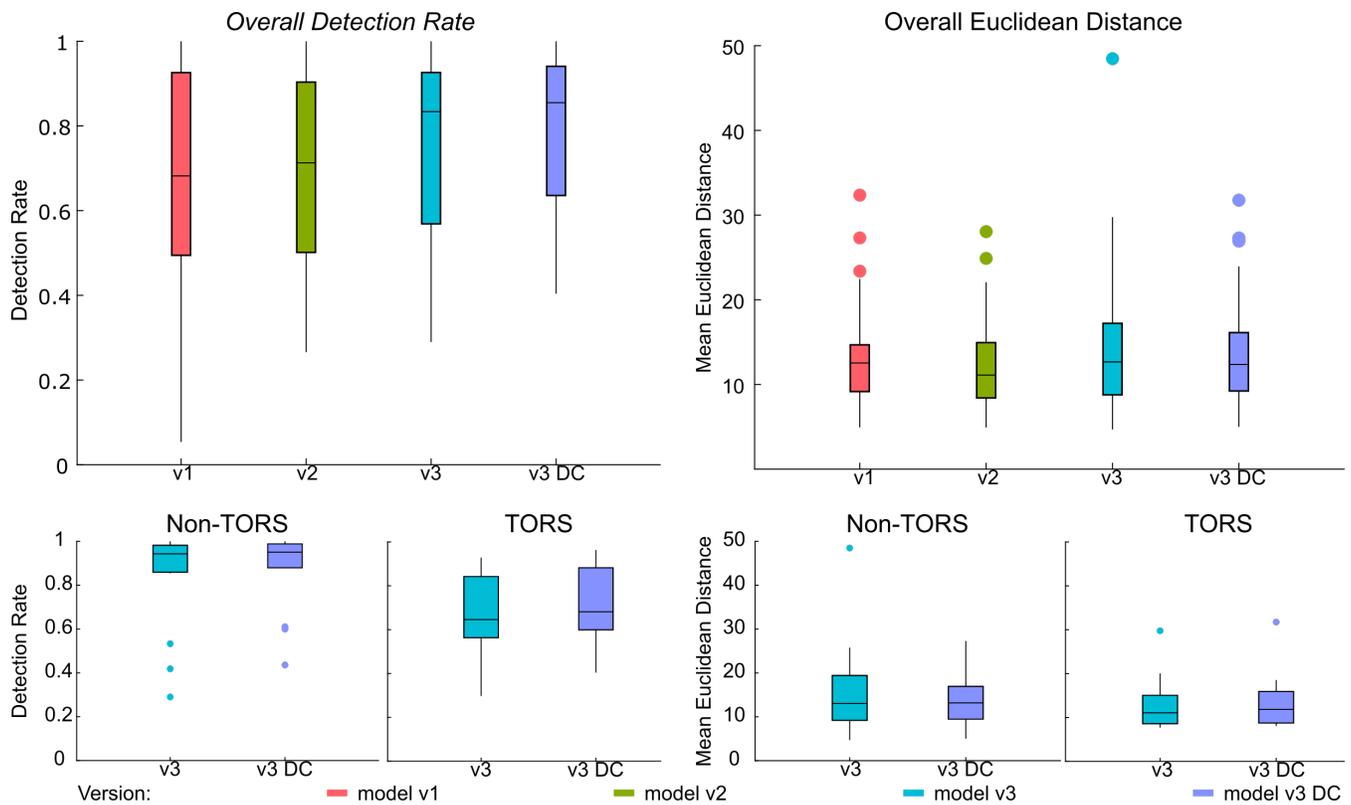

Figure 5: Comparative analysis of the distribution of Detection Rate and Mean Euclidean Distance across different models. The upper panel shows the overall detection rate and precision in localizing the aiming beam point on the test dataset, which consists of 40 in vivo FLIm scan videos recorded during head and neck surgical procedures. The lower panel presents the comparative performance of the models by categorizing the test dataset into two types of surgical procedures: TORS (24 videos) and non-TORS (16 videos). The data highlights the relative performance of each model in terms of detection rate and consistency in localizing the aiming beam across different surgical context.

In terms of precision, as measured by the Mean Euclidean Distance, Model v2 demonstrated superior performance, with a mean value of 12.41 pixels, a standard deviation of 4.98 pixels, and a median value of 11.11 pixels (Fig. 5). Although Model v3 DC exhibited a slightly higher mean Euclidean distance of 13.45 pixels, it maintained consistent performance, as evidenced by the high Pearson correlation coefficients for the 'x' and 'y' coordinates across models (Table. III). It is important to note that the Euclidean distance and Pearson correlation coefficients were computed based on the subset of aiming beam points detected by each model. The consistency of these metrics underscores the precision of aiming beam localization when the aiming beam is successfully detected.

### A. Detection Rate vs. Surgery Type

In addition to the overall evaluation, we further analyzed model performance by categorizing the dataset into two surgery types TORS and non-TORS procedures. The detection rate comparison between Model v3 and Model v3 DC for these two categories is illustrated in Fig. 5. Overall, Model v3 DC consistently outperformed Model v3 in both TORS and Non-TORS procedures. However, we observed a significant decrease in detection rate during TORS surgery compared to non-TORS procedures. Specifically, Model v3 DC achieved a mean detection rate of 0.71 for TORS, compared to a higher mean detection rate of 0.88 for non-TORS. Despite the variation in detection rates, the precision of aiming beam point localization remained consistent across both surgery types.

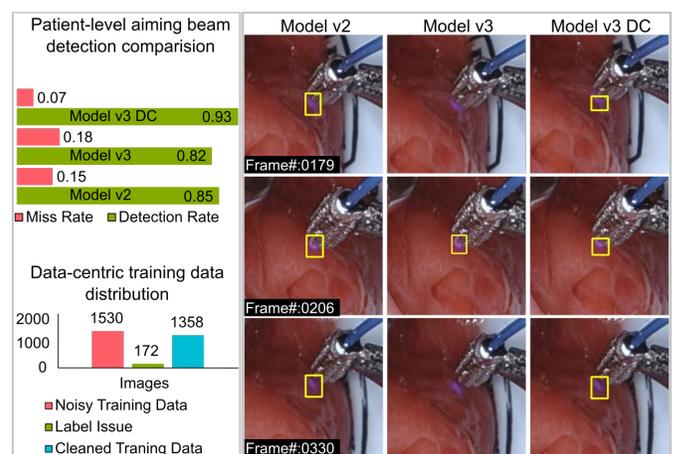

Figure 6: Patient-level analysis of aiming beam detection highlights the impact of the data-centric (DC) learning approach. Left panel upper compares the performance of three models—Model v2, Model v3, and Model v3 DC—in detecting the aiming beam for patient 'A'. left panel lower illustrates the training data distribution and the label issues identified after processing the dataset with the data-centric learning approach. Right panel shows a sequence of frames (Frame# 0179, 0206, and 0330) from patient 'A', demonstrating the detection of the aiming beam (highlighted by the yellow box) during a surgical procedure, using the three models.

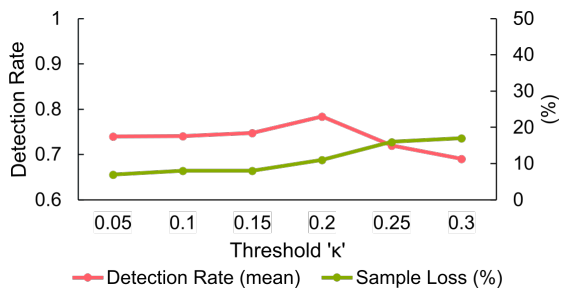

Figure 7: Performance evaluation of the aiming beam detection rate for model v3 DC as a function of the confidence threshold ($\kappa$). The results indicates the threshold impacts the aiming beam detection rate, with the best performance observed around a threshold of 0.2.

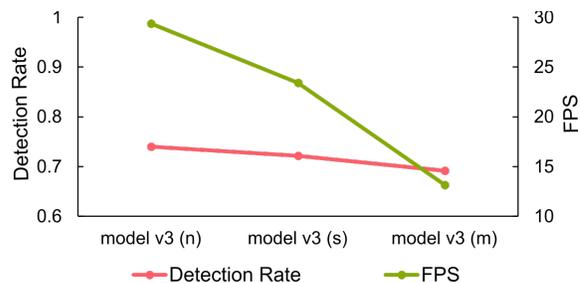

Figure 8: Comparison of the aiming beam detection rate and computational efficiency for the YOLOv8 models: nano (n), small (s), and medium (m). The results indicate that the nano version offers an optimal balance between detection accuracy and computational efficiency.

### B. Patient-level Model Robustness

In evaluating the robustness of the models at the patient level, we selected a representative patient who exhibited inconsistencies observed across multiple cases to illustrate this variability (see Fig. 6). We analyzed the aiming beam detection rate across different models for this patient. Notably, Model v2, trained with 60% of the available data, achieved a detection rate of 0.85. Surprisingly, Model v3, which was trained with 100% of the data, exhibited a lower detection rate of 0.82, failing to detect aiming beam points that were successfully identified by the previous model version (see Fig. 6(a) and Fig. 6(b)). This unexpected result suggests that simply increasing the size of the training dataset does not automatically enhance model performance and consistency for specific patient. On the contrary, it may introduce new challenges, such as change to the distribution of training data or learning from incorrect annotations, which can ultimately hinder the model's ability to generalize effectively.

However, introducing a data-centric approach in Model v3 DC led to a significant improvement, with the detection rate for the patient rising to 0.93 and the miss rate dropping to 0.07 (see Fig. 6(a)). This model not only retained the successful detections from Model v2 but also recovered the missed detections from Model v3. The data-centric methodology employed in Model v3 DC effectively mitigated the issues introduced by the larger dataset, enhancing the model's robustness and reliability at the patient level. This improvement is particularly critical in the clinical translation domain, where visualization methods must perform consistently across new patients.

### C. Impact of Threshold '$\kappa$' on Detection Rate

The threshold '$\kappa$' plays a crucial role in our data-centric methodology, as it determines which samples are included in the training dataset based on their confidence scores. We evaluated the impact of varying $\kappa$ on the detection rate, and the results are shown in Fig. 7. The analysis shows that as the threshold $\kappa$ increases from 0.05 to 0.2, the mean detection rate gradually improves, reaching its peak at a detection rate of 0.784 when $\kappa$=0.2. Beyond this point, increasing $\kappa$ further results in a decline in the detection rate, with a significant drop observed at $\kappa$=0.3. Based on these findings, we selected a threshold of $\kappa$=0.2, which resulted in a 9% reduction in training samples. This value strikes an optimal balance, filtering out noisy or uncertain samples while retaining enough high-confidence data to train a robust model. The decline in performance at higher thresholds indicates that overly stringent filtering may exclude too much data, leading to a less effective model. Thus, the choice of $\kappa$=0.2 maximizes the detection rate and ensures a well-balanced trade-off between data quality and quantity.

### D. Evaluating Yolov8 model versions and Computation Efficiency

We assessed the performance of three versions of the YOLOv8 model—n (nano), s (small), and m (medium)—focusing on both detection rate and computation speed, measured in frames per second (FPS), as shown in Fig. 8. The evaluation revealed that Model v3 (n), the smallest version, achieved the highest mean detection rate of 0.740, along with the fastest computation speed, averaging 29.37 FPS.

This analysis highlights the importance of selecting the appropriate model version based on the specific requirements of the application. Although larger models like Model v3 (m) may theoretically handle more complex data better, they are less suitable for practical deployment due to their slower processing speed. Additionally, the reduced detection accuracy of the larger models '*m*' and '*s*' could be attributed to the limited size of the training dataset, suggesting that these models may require more extensive data to fully realize their potential.

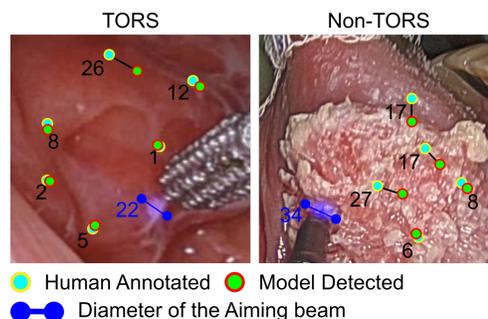

Figure 9: Illustrates the distribution of Euclidean distances in pixels between the detected aiming beam points (green with red outline) and the annotated points (turquoise with yellow outline) for a patient undergoing TORS and non-TORS procedures. The illustration of the Euclidean distance offers insights into the precision of aiming beam point localization and the variation in aiming beam diameter across scans, as indicated by the purple annotations.

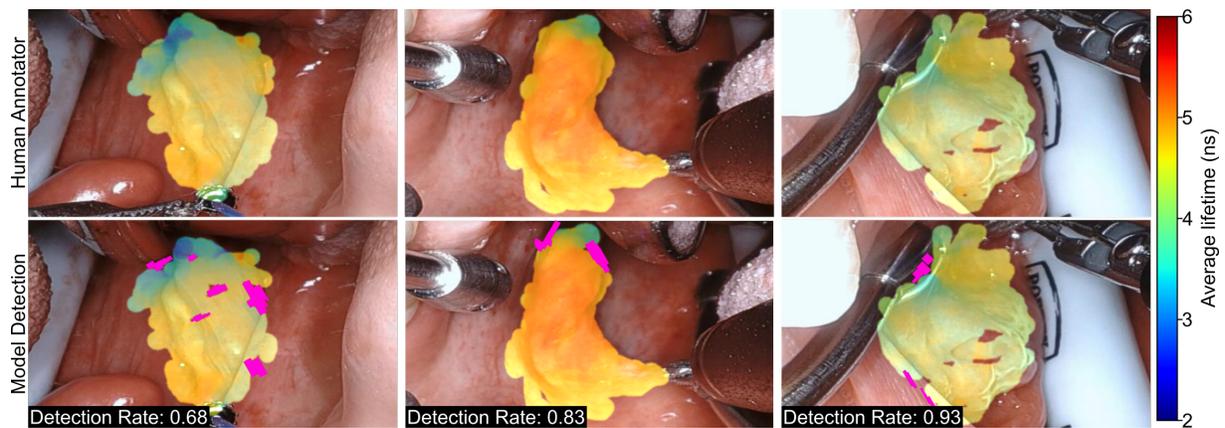

Figure 10: The scan coverage visualization in relation to the detection rate. Here we augment the average lifetime from channel 2 of the FLIm system. The upper panel illustrates the augmented overlay from the human annotated aiming beam points. The lower panel illustrates overlay visualization from the aiming beam point localization from the detected aiming beam of model v3 DC. The 'magenta' annotations highlighting the missed coverage of the overlay visualization.

### E. Real-time Model Performance

The performance of the deployed model (i.e., Model v3) in the clinical system was assessed for four FLIm scans during TORS procedure. When compared against human annotator, the model reported a detection rate with a mean of 0.867, a median of 0.855, and a standard deviation of 0.086. The overall computational efficiency of the system, including the overlay visualization pipeline [9] was approximately 24 FPS, demonstrating its suitability for real-time surgical guidance.

### F. Model Performance in Context of Head and Neck Surgery

The model v3 DC reported a mean Euclidean distance of 13.44 pixels, with a standard deviation of 5.93 pixels, when comparing the aiming beam localization coordinates to the human-annotated reference points (Fig. 9). However, the precision of localization in pixel units does not fully capture the method's practical effectiveness in a surgical context. Since surgeons evaluate the reliability of the system based on real-world dimensions rather than pixel measurements, we employed a pixel-to-millimeter conversion using a known reference object in the scene, such as the FLIm probe. By leveraging the probe's known physical dimensions and the corresponding pixel measurements in the instance segmentation mask, a pixel range of 10 to 40 pixels corresponded to approximately 0.5 mm to 2.6 mm in the real world. The mean Euclidean distance of 13.44 pixels translates to the lower end of this range, which is within acceptable limits considering that surgeons maintain a distance of greater than 5 mm from invasive tumor to resected margin to achieve a clear margin [21], [22].

The scan coverage visualization in relation to the detection rate is illustrated in Fig. 10, with the upper panel representing the human-annotated aiming beam and the lower panel showing the model-detected aiming beam. The comparison indicates that as the detection rate improves—from 0.68 to 0.93—the model's visualization increasingly aligns with the human annotations. At lower detection rates, the overlay tends to have incomplete regions (highlighted by the black annotations), primarily along the outer boundary and smaller gaps near the center of the scan. Conversely, at higher detection rates, the model's output closely matches the human annotations, highlighting the critical importance of achieving high detection rates to achieve consistent scan coverage, which is essential for effective surgical guidance.

## IV. DISCUSSION

This study presents a deep learning-based approach for real-time detection of aiming beam from fiber-based FLIm scans during head and neck surgery. The proposed data-centric methodology significantly improves the accuracy of aiming beam detection, with higher detection rates and precision across different surgical scenarios.

### A. Impact of Data Centric Learning

Incorporating the data-centric (DC) approach into our pipeline significantly enhanced the model's generalization and robustness. By addressing label issues and training on cleaner data, we reduced noise, which minimized the risk of the model learning spurious correlations. This enabled the model to focus on true underlying patterns, improving its performance across unseen surgical environments and patient variations. With cleaner data, the model achieved more consistent detection accuracy in both TORS and non-TORS procedures. Additionally, by training on high-confidence, accurately labeled examples, we further reduced the likelihood of overfitting, resulting in more reliable predictions when deployed in varied and challenging clinical settings.

Although there are trade-offs, such as the potential loss of data diversity due to the pruning process, the benefits of the data-centric approach are evident. Approximately 9% of the training data was removed due to label issues identified through the DC process. A key issue, illustrated in Fig. 4, involves reflections of the aiming beam on adjacent tissues. Although these reflections were correctly not labeled as the aiming beam by the annotator, as they should be considered background, the model still flagged this data as problematic. This indicates that the model may struggle to distinguish between the actual aiming beam and its reflections, potentially causing confusion during training. Other label issues include cases where the aiming beam is barely visible and difficult to detect in the surgical view, as well as instances where it is completely absent. In contrast, the clean dataset consists of examples where the aiming beam is clearly visible in the surgical view, representing high-quality data that is ideal for training. This distinction between clean and problematic data underscores the importance of careful data curation in improving model performance, despite the modest reduction in data size. These findings emphasize the critical role of clean and representative training data in developing reliable models for surgical guidance

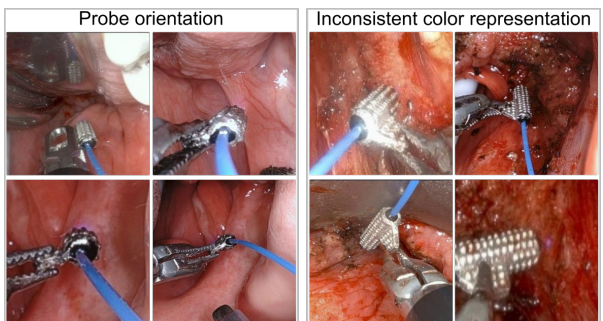

Figure 11: Challenges in aiming beam detection during TORS procedures due to the lack of a visible beam within the scene. The left panel illustrates probe orientation issues, particularly when the probe is nearly perpendicular to the field of view of the endoscopic camera. The right panel highlights the inconsistent color representation in the field of view.

### B. Challenges in Detecting Aiming Beam

We observed a notable difference in detection rates between TORS and non-TORS procedures. TORS, a minimally invasive robotic surgical procedure, involves navigating the fiber probe through constrained spaces, particularly around the base of the tongue and palatine tonsil regions. The complex anatomy and restricted access make it challenging for surgeons to maneuver the probe so that the aiming beam remains visible within the surgical view. As shown in Fig. 11, detecting the aiming beam becomes particularly challenging when the probe is nearly perpendicular to the field of view of the TORS endoscopic camera. Additionally, inconsistent color representation within the field of view further impacts detection, often resulting in a lack of a visible beam within the scene.

In contrast, non-TORS procedures typically involve the use of a handheld probe to scan the oral cavity, presenting fewer challenges for the surgeon in terms of probe orientation. This ease of maneuverability leads to a significantly higher detection rate in non-TORS procedures, as the aiming beam is more consistently visible within the surgical field. However, despite the differences in detection rates between TORS and non-TORS procedures, it is important to note that when the aiming beam is detected, the detection is generally precise, regardless of the procedure type.

### C. Latency Requirements for Real-Time Surgical Guidance

In real-time surgical guidance, achieving low latency is crucial for maintaining precise and responsive control, as it directly impacts a surgeon's ability to interact seamlessly with augmented overlays or robotic instruments. Studies examining robotic surgeries have shown that communication delays beyond 100 ms can impair surgical control, disrupting task completion times, reduce precision, and increase unintended movements [23], [24], [25]. For complex, high-precision tasks, recent studies recommend keeping latency under 70 ms to ensure optimal performance [26]. Our system, with an approximate latency of 41.7 ms per frame at 24 FPS, operates well within this threshold, allowing for seamless aiming beam detection and FLIm overlay augmentation. This responsiveness is vital for effective intraoperative guidance.

## V. Conclusion

This study presents a data-centric approach to enhance real-time surgical guidance using fiber-based FLIm by accurately detecting the aiming beam in complex surgical environments. By employing a data-centric learning strategy that prunes label noise, we achieved substantial improvements in detection accuracy. The model consistently performed with a median detection rate of 85% across both in vivo surgical videos and clinical system deployment, maintaining computational efficiency suitable for real-time applications. These results underscore the critical role of data-centric learning in developing robust deep learning models for surgical guidance. Future efforts will focus on further enhancing the model's detection rate and broadening its application to diverse surgical procedures, facilitating its widespread clinical adoption.

## Acknowledgment

The authors thank our clinical research coordinator Randev Sandhu, CCRP at the University of California, Davis Medical Center Department of Otolaryngology for their many contributions to enroll and consent patients in our study, maintain patient files and medical history documents, and disseminate research study information to our team. We would like to acknowledge Dr. Jonathan Sorger (Intuitive Surgical, Sunnyvale CA) for his support for our ongoing industry collaboration; key areas of his industry support include FLIm visualization aspects and integration of FLIm fiber optic probes into the da Vinci SP TORS platform. Finally, we are grateful to our lab members Divyanshu Malik, Vikram Parimal Karmarkar, Suparn Sathya, and Nabeel Sabzwari for participating in the data annotation of the training dataset.